\documentclass[10pt,twocolumn,letterpaper]{article}

\usepackage[algorithms]{wacv}      

\usepackage{graphicx}
\usepackage{amsmath}
\usepackage{amssymb}
\usepackage{booktabs}
\usepackage{multirow}
\usepackage[table,xcdraw]{xcolor}

\usepackage{soul}
\usepackage{bbm}
\usepackage{xcolor}
\usepackage[symbol]{footmisc}

\newcommand\numberthis{\addtocounter{equation}{1}\tag{\theequation}}

\newcommand{\best}[1]{\textbf{\underline{#1}}}
\newcommand{\second}[1]{\underline{#1}}

\usepackage{pifont}
\newcommand{\cmark}{\ding{51}}%
\newcommand{\xmark}{\ding{55}}%
\newcommand{\ours}{XCNorm}
\usepackage[pagebackref,breaklinks,colorlinks]{hyperref}

\usepackage[capitalize]{cleveref}
\crefname{section}{Sec.}{Secs.}
\Crefname{section}{Section}{Sections}
\Crefname{table}{Table}{Tables}
\crefname{table}{Tab.}{Tabs.}

\def\wacvPaperID{191} 
\def\confName{WACV}
\def\confYear{2024}

\begin{document}

\title{Single Domain Generalization via Normalised Cross-correlation Based Convolutions}

\author{WeiQin Chuah\footnotemark[1] \hspace{1.5ex} Ruwan Tennakoon\footnotemark[1] \hspace{1.5ex} Reza Hoseinnezhad\footnotemark[1] \hspace{1.5ex} David Suter\footnotemark[2] \hspace{1.5ex} Alireza~Bab-Hadiashar\footnotemark[1]\\
RMIT University, Australia\footnotemark[1] \hspace{3.5ex} Edith Cowan University~(ECU), Australia\footnotemark[2]\\
{\tt\small \{wei.qin.chuah,ruwan.tennakoon,rezah,abh\}@rmit.edu.au, d.suter@ecu.edu.au}}
\maketitle

\begin{abstract}
Deep learning techniques often perform poorly in the presence of domain shift, where the test data follows a different distribution than the training data. The most practically desirable approach to address this issue is Single Domain Generalization (S-DG), which aims to train robust models using data from a single source. Prior work on S-DG has primarily focused on using data augmentation techniques to generate diverse training data. In this paper, we explore an alternative approach by investigating the robustness of linear operators, such as convolution and dense layers commonly used in deep learning. We propose a novel operator called ``\textit{\ours}''~that computes the normalized cross-correlation between weights and an input feature patch. This approach is invariant to both affine shifts and changes in energy within a local feature patch and eliminates the need for commonly used non-linear activation functions. We show that deep neural networks composed of this operator are robust to common semantic distribution shifts. 
Furthermore, our empirical results on single-domain generalization benchmarks demonstrate that our proposed technique performs comparably to the state-of-the-art methods.
\end{abstract}

\section{Introduction}
\label{sec:intro}

\begin{figure}
    \centering
    \includegraphics[width=0.38\textwidth]{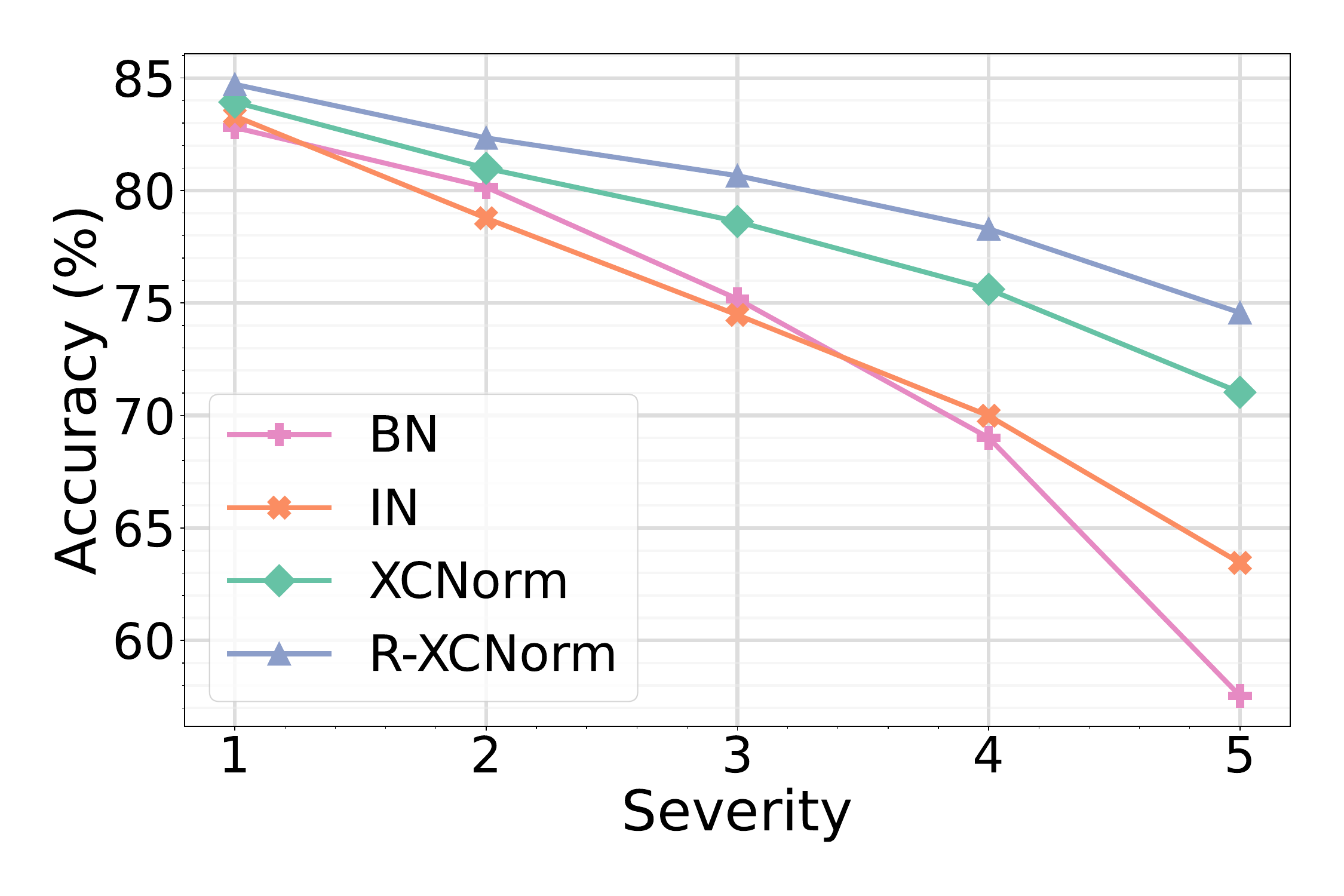}
    \caption{Comparison of accuracy for common normalization methods in image classification on CIFAR-10-C, considering five levels of domain discrepancy caused by corruptions. The evaluated methods include BN, IN, and our proposed approaches (\ours~and R-\ours). }
    \label{fig:norm_methods}
\end{figure}

Deep learning techniques have achieved practical success in a variety of fields, including computer vision, natural language processing, and speech processing. However, this success is often limited to settings where the test data follows the same distribution as the training data. 
In many real-world situations, this assumption breaks down due to shifts in data distribution, known as \textit{domain-shift}~\cite{ben2010theory}, which can significantly degrade performance \cite{taori2020measuring}. 

Dealing with domain-shift is a challenging problem with important practical implications. There are two main approaches to address domain shift and enable the transfer of knowledge from previously seen environments (\textit{source} domains) to a new environment (\textit{target} domain) without using any labeled data of the target domain: 
(1) Domain Adaptation~\cite{wilson2020survey} (DA) where a model trained with source data is recalibrated using unlabeled data from the target domain, and (2) Domain generalisation~\cite{zhou2022domain} (DG) where a model is trained on multiple source domains but no target domain data is available for recalibration.
The most \textit{data-efficient} domain generalisation technique is
the single domain generalisation (S-DG), which requires data from only a single source domain to train a model that is robust against unforeseen data shifts. Although practically desirable, S-DG has received little attention in the past.



S-DG presents a significant challenge due to two main factors. Firstly, the input data, derived from only one source domain, does not provide sufficient opportunity to observe the possible diversity in out-of-domain data. Secondly, the presence of spurious correlations or shortcuts can further complicate the issue by introducing biases and hindering generalization. 
Prior work on S-DG has primarily focused on increasing the diversity of input data using adaptive data augmentation techniques. 
These include creating fictitious examples that mimic anticipated shifts in data distribution using random~\cite{xu2021robust}, adversarial~\cite{volpi2018generalizing, qiao2020learning, zhao2020maximum, fan2021adversarially} or causality~\cite{chen2023meta, ouyang2022causality} based data augmentation, as well as image style diversification~\cite{wang2021learning}.


The generalization of a model is largely influenced by its support, which refers to the diversity of its input data and its inductive biases~\cite{wilson2020bayesian}. While not explicitly stated as such, the success of the above-mentioned S-DG methods hinges on increasing the input diversity via data augmentation. An alternative and complementary approach that has received less attention is to incorporate inductive biases into the network components to make them more robust to domain shifts. In this paper, we explore the above approach and propose a robust alternative to linear operators, such as convolution and dense layers, which are fundamental components of most neural networks.

We draw on the classical idea of template matching and consider linear layers in a neural network as computing a matching between the template (represented by the weights) and the signal (represented by input feature maps) using \textit{cross-correlation}, as detailed in \Cref{equ:typical_conv}.  
Early works in template matching have shown that cross-correlation is not ideal for pattern matching as it fails when the local energy of the input (i.e., $\sum_{u,v} z[u,v]^2$) varies, and is not robust to affine shifts in the input signal~\cite{lewis1995fast}. More recently, Jin~\etal~\cite{jin2022domain} empirically showed that domain shift primarily causes variation in the local energy of feature representations. This suggests that the linear operator, which is sensitive to local energy, might degrade out-of-domain (OOD) generalization.

The above perspective enables us to use more robust template-matching techniques such as normalized cross-correlation~\cite{lewis1995fast} to replace convolutions or dense layers in neural networks and recover the underlying invariant features in the input.
We call our method ``\textit{\ours}'', which performs cross-correlation between standardized (i.e., Z-score normalized) weights and patch-wise standardized input feature maps. 
This reduces the influence of input feature magnitude on the output and makes the operator invariant to affine transformations of the input. Moreover, we leverage robust statistics to improve the resilience of \ours~to outliers and introduce a refined version of our method named \textit{R-\ours}. 
As \Cref{fig:norm_methods} demonstrates, our methods achieve significantly better robustness to semantic distribution shifts on CIFAR-10-C, in contrast to other normalization techniques. Moreover, the advantage of our methods becomes more pronounced as the domain discrepancy increases.
The contributions of this paper include:
\begin{itemize}
    \item We propose a novel nonlinear operator called ``\textit{\ours}'', based on normalized cross-correlation, that reduces the influence of input feature magnitude on the output and invariant to affine transformations. 
    \item Leveraging robust statistics, we further enhance the robustness of \ours~to outliers. Our experiments on several commonly used benchmarks in S-DG show that the proposed robust operator (``\textit{R-\ours}'') is also complementary to augmentation-based methods and achieves state-of-the-art performance.
    \item We empirically show that a neural network based on ``\textit{\ours}'' or ``\textit{R-\ours}'', is significantly more robust to semantic shifts compared to a network based on a typical linear operator.   
\end{itemize}





\section{Related Work}

\noindent \textbf{Domain Generalization:}
Domain generalization (DG) methods aim to learn robust models from several source domains that can generalize to unseen target domains, addressing the issue of domain shifts. 

A particularly challenging yet practical variant of domain generalization (DG) is a single-domain generalization (S-DG), where only one source domain is available during training. S-DG is especially challenging because, unlike in DG, there is no access to multiple source domains that would allow for the observation of possible shifts in data and invariances between domains. To address this challenge, researchers have primarily focused on using data augmentation techniques to generate diverse training data and increase input diversity. 
A common technique is posing S-DG as a ``distributionally robust optimization'' problem and solving it using adversarial data augmentation (ADA)~\cite{volpi2018generalizing}. ADA lacks the ability to produce large semantic shifts that are common in real data. As a result, subsequent works have added additional constraints to adversarial augmentation~\cite{qiao2020learning, zhao2020maximum, li2021progressive, fan2021adversarially} or incorporated background knowledge about anticipated semantic shifts via random augmentations~\cite{xu2021robust}, causality based data augmentations~\cite{chen2023meta, ouyang2022causality}, or image style diversification~\cite{wang2021learning}.

An alternative that has received little attention is to incorporate inductive biases into the network components to make them more robust to domain shifts. The most closely related work in this direction is the Meta Convolution Neural Networks~(Meta-CNN) proposed by Wan~\etal~\cite{wan2022meta}, where the output feature-maps of each layer are reconstructed using templates learned from training data, resulting in universally coded images without biased information from unseen domains. Our proposed operator, \textit{\ours}, offers a simpler implementation compared to \cite{wan2022meta}. While their method involves more complex operations, our approach simply replaces the convolution function with our \ours~method. This simplicity makes our method more straightforward to implement and integrate into existing frameworks.


\vspace{.5em}
\noindent \textbf{Normalization in Neural Networks:} During the training process of a deep neural network, the input distribution of an intermediate layer continuously changes, a phenomenon known as covariate shift. This makes it challenging to set the hyper-parameters of a layer, such as the learning rate. To tackle this issue, various normalization techniques have been proposed, including Batch Norm~\cite{ioffe2015batch}, Instance Norm~\cite{ulyanov2016instance}, GroupNorm~\cite{wu2018group}, and Layer Norm~\cite{ba2016layer}, which aim to normalize the output of each layer using batch statistics, feature channels, groups of features, or the entire layer's output, respectively. Instead of operating on features, Weight Norm~\cite{salimans2016weight} proposes normalising the filter weights.



While most of the aforementioned work has focused on in-domain generalization, there are a few studies that have examined generalization ability under domain shift. For instance, BN-Test~\cite{nado2020evaluating} computed batch normalization statistics on the test batch while  DSON~\cite{seo2020learning} used multi-source training data to compute the statistics.
Fan~\etal~\cite{fan2021adversarially} investigated normalization for single-domain generalization, where adaptive normalization statistics for each individual input are learned. These adaptive statistics are learned by optimizing a robust objective with adversarial data augmentation.
The above works view normalization as being independent of the base operator (e.g., convolution, fully-connected). In contrast, our approach considers normalization to be an integral part of the base operator. We normalize both weights and input for each local spatial region of the input.

\vspace{.5em}
\noindent \textbf{Non-linear Transforms:} Several works have explored the use of non-linear transforms to replace the linear transform in the typical convolution operator~\cite{ghiasi2019generalizing, liu2018decoupled, luo2018cosine, liu2017deep, wang2019kervolutional, zoumpourlis2017non}. The works most closely related to ours are those by  \cite{liu2018decoupled, luo2018cosine, liu2017deep, bohle2022b} assess the cosine similarity between weights and inputs to improve both model performance and interpretability. In those methods, the convolution is viewed as an inner product between an input patch $\mathbf{z}$ and the weights $\mathbf{w}$:
\begin{align}
    \mathrm{Conv}(\mathbf{z}; \mathbf{w}) =  \left< \mathbf{z}, \mathbf{w} \right> &{} = \left\| \mathbf{z} \right\| \left\| \mathbf{w} \right\| \cos{(\varphi)}\\
    &{} = h\left ( \left\| \mathbf{z} \right\|, \left\| \mathbf{w} \right\| \right) g \left( \cos{(\varphi)} \right)
\end{align}
where $\cos{(\varphi)}$ is the angle between $\mathbf{z}$ and $\mathbf{w}$. This view allows for the separation of norm terms from the angle term (decoupling), and to change the form of $h()$ and $g()$ independently. Liu \etal~\cite{liu2018decoupled} derived several decoupled variants of the functions $h()$ and $g()$. They demonstrated that the decoupled reparameterizations lead to significant performance gains, easier convergence, and stronger adversarial robustness. Later \cite{bohle2022b} introduced the ``B-cos'' operator and showed that it lead to better neural network interpretations. 

Our proposed \textit{\ours} operator can also be seen within this framework, where the dot product is taken between the centered and normalized versions of the input patch and the weights. 
However, unlike the methods mentioned above, we investigate the use of \textit{\ours} for out-of-domain generalization in a single source domain setting.

\section{Method}


Given a source domain $\mathcal{S} = \left\{ \left( x^\mathcal{S}_i, y^\mathcal{S}_i \right)\right\}_{i=1}^{N_\mathcal{S}} \sim P_{XY}^{\mathcal{S}}$ the goal of Single Domain Generalization is to learn a robust and generalizable predictive function $f_\theta : \mathcal{X} \rightarrow \mathcal{Y}$ that can achieve a minimum prediction error on an unseen target domain $\mathcal{T} \sim P_{XY}^{\mathcal{T}}$. Here, the joint distribution between the domains is different i.e.  $P_{XY}^{\mathcal{S}} \neq P_{XY}^{\mathcal{T}} $.

\begin{figure}
    \centering
    \includegraphics[width=.2\textwidth]{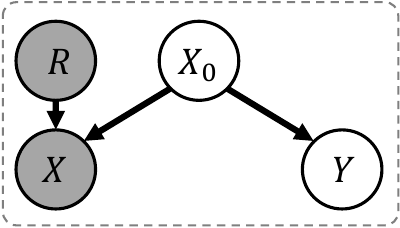}
    \caption{Probabilistic graphical model representing the data generation process. Circles denote random variables, such as $X$ as input (covariates) and $Y$ as target variable, and solid arrows represent direct dependencies between variables. The shaded circles $R$ and $X$ indicate that the distribution of those random variables shifts between source and target environments.}
    \label{fig:dataset_shift}
\end{figure}

Ben-David~\etal~\cite{david2010impossibility} showed that it would not be possible to learn models that can generalize to any distribution beyond the source distribution using solely the data sampled from that source distribution. Therefore, to generalize, it is essential to impose restrictions on the relationship between the source and target distributions. One common assumption in S-DG is that the target variable $Y$ depends on an underlying latent representation in the covariate space $X_{0}$, that remains invariant across domains. However, $X_{0}$ cannot be directly observed; instead, we observe a mapping of $X_{0}$ into the observable space $X$, controlled by decision attributes $R$ such as rendering (synthetic data) or data capture (real data) parameters. These attributes often change between the source and target domains, causing domain shifts. This assumption is represented by a Probabilistic graphical model, which is shown in \Cref{fig:dataset_shift}.

Most S-DG methods based on data augmentation aim to diversify $X$ so that it spans the range of possible $R$ values~\cite{xu2021robust, volpi2018generalizing, qiao2020learning, zhao2020maximum, fan2021adversarially, chen2023meta, ouyang2022causality, wang2021learning}. In contrast, our approach is to modify the model to make it robust to variations caused by $R$. For this purpose, we draw on the classical idea of template matching and consider linear layers in neural networks as computing a matching between the template (represented by the weights) and the signal (represented by input feature maps).  This perspective enables us to use more robust template-matching techniques such as normalized cross-correlation to replace convolutions or dense layers in neural networks and recover the underlying invariant features $X_{0}$ in the input.


\subsection{Normalized Cross-Correlation Layer}
Typically, the linear units of a DNN layer compute the cross-correlation between the input feature maps $\mathbf{z} \in \mathbb{R}^{H \times W \times C_{in}}$~\footnote[1]{Batch dimension is omitted for simplicity.} and the weights $\mathbf{w} \in \mathbb{R}^{K \times K \times C_{in} \times C_{out} }$~\footnote[2]{We use notations that are consistent with convolutional layers for convenience. For a fully connected (dense) layer, we assume that $H = W = K = 1$.}. Here $C_{in(out)}$ is the number of channels in the input (or output), $[H, W]$ are the feature map spatial dimensions and $K$ is the kernel size. The pixel $(u,v)$ of the $c^\mathrm{th}$ output feature channel is computed as:
\begin{equation}
     \Phi_{u,v,c}(\mathbf{z}; \mathbf{w}) = \left< \mathbf{z}_{ \Tilde{u},\Tilde{v}} , \mathbf{w}_c \right> = \sum_j {z}_{\Tilde{u},\Tilde{v}}^{(j)} \cdot {{w}}_c^{(j)} 
     \label{equ:typical_conv}
\end{equation}
Here $\mathbf{z}_{ \Tilde{u},\Tilde{v}} $ is a patch of the input feature map centered at $ \Tilde{u},\Tilde{v}$ with the same shape as the weight tensor $\mathbf{w}_c$, and $j$ index the pixels within the patch. The map $\nu: (u, v) \rightarrow (\Tilde{u},\Tilde{v}) $ is determined by parameters of the convolution (or dense) layer (i.e., stride, kernel width). 

The operator above is ineffective for pattern matching when the patch energy (i.e., $\left \| \mathbf{z}_{ \Tilde{u},\Tilde{v}} \right \| = \sum_{j} ({z}_{\Tilde{u},\Tilde{v}}^{(j)})^2$) is not uniform across a feature map. It also lacks robustness to affine transformations of the input, i.e., $\left< \mathbf{z}_{ \Tilde{u},\Tilde{v}} , \mathbf{w}_c \right> \neq \left< \mathbf{A} \mathbf{z}_{ \Tilde{u},\Tilde{v}} + \mathbf{b} , \mathbf{w}_c \right>$~\cite{lewis1995fast}.
To overcome the above limitations, we propose using the normalized cross-correlation operator as a replacement for the linear operator. With this new operator, which we coin \textit{\ours}, the output feature at position $(u,v)$ of the $c^\mathrm{th}$ feature channel is computed as:
\begin{align*}
     \Psi_{u,v,c}(\mathbf{z}; \mathbf{w}) 
     &{}= \frac{\sum_j [ {z}_{\Tilde{u},\Tilde{v}}^{(j)} - \mu_{\mathbf{z}(\Tilde{u},\Tilde{v})} ]  [ {w}^{(j)}_c - \mu_{\mathbf{w}_c} ]}{    \left \| \mathbf{z}_{\Tilde{u},\Tilde{v}} - \mu_{\mathbf{z}(\Tilde{u},\Tilde{v})} \right \|  \left \| \mathbf{w}_c - \mu_{\mathbf{w}_c} \right \| }\numberthis \label{eqn:ncc_first1} \\
     &{}= \frac{\sum_j {z}_{\Tilde{u},\Tilde{v}}^{(j)}  \cdot  {w}^{(j)}_c - \alpha \mu_{\mathbf{z}(\Tilde{u},\Tilde{v})} \mu_{\mathbf{w}_c}}{ {\alpha}\sigma_{\mathbf{w}_c}  \sqrt{  \frac{1}{\alpha}\sum_j {z}_{\Tilde{u},\Tilde{v}}^{(j)}  \cdot {z}_{\Tilde{u},\Tilde{v}}^{(j)} -  \mu_{\mathbf{z}(\Tilde{u},\Tilde{v})}^2 } + \epsilon}.
     \numberthis \label{eqn:ncc_first}
\end{align*}
Here, $\mu_{\bullet}$ is the mean of $\bullet$, $  \sigma_{\mathbf{w}_c} =  \sqrt{\frac{1}{\alpha}\sum_j [ {w}^{(j)}_c - \mu_{\mathbf{w}_c} ]^2}$, $\epsilon$ is a small constant to ensure numerical stability, and $\alpha = K \times K \times C_{in}$ is the number of pixels in the patch. 

Since the patch-wise mean, $\mu_{\bullet}$, can be computed using linear operations (i.e., convolving with constant weight tensor with all elements equal to $1/\alpha$), the \ours~can be realized using linear operators:
\begin{align}
    \Psi(\mathbf{z}; \mathbf{w}) 
    {}& = \frac{\Phi(\mathbf{z}; \mathbf{w}) - \alpha ~\textcolor{blue}{\Phi(\mathbf{z}; \mathbf{w}_{\alpha)}} ~\boldsymbol{\mu}_{\mathbf{w}} }{ \alpha~\sqrt{\left [  \textcolor{violet}{\Phi(\mathbf{z}^2 ; \mathbf{w}_\alpha)} - \textcolor{blue}{\Phi(\mathbf{z} ; \mathbf{w}_\alpha)^2} \right]  } ~ \boldsymbol{\sigma}_{\mathbf{w}} + \epsilon }\\
    {}& = \frac{\Phi(\mathbf{z}; \mathbf{w}) - \alpha~\textcolor{blue}{\boldsymbol{\mu}_{\mathbf{z}}} ~ \boldsymbol{\mu}_{\mathbf{w}} }{ \alpha~\sqrt{ \left [  \textcolor{violet}{ \boldsymbol{\mu}_{\mathbf{z}^2}  } - \textcolor{blue}{\boldsymbol{\mu}_{\mathbf{z}}^2} \right ]}~ \boldsymbol{\sigma}_{\mathbf{w}}  + \epsilon}.
\end{align}
Here, $\mathbf{w}_\alpha \in \mathbb{R}^{K \times K \times C_{in} \times 1}$ is a constant matrix with all elements equal to $1/\alpha$, $\boldsymbol{\mu}_{\mathbf{w}} \in \mathbb{R}^{1 \times C_{out} }$ is the mean of the weights within each output channel, and $\boldsymbol{\sigma}_{\mathbf{w}} \in \mathbb{R}^{1 \times C_{out} }$ is the variance of the weights within each output channel. 

\begin{figure}[t]
     \centering
     \begin{subfigure}[b]{0.23\textwidth}
         \centering
         \includegraphics[width=\textwidth]{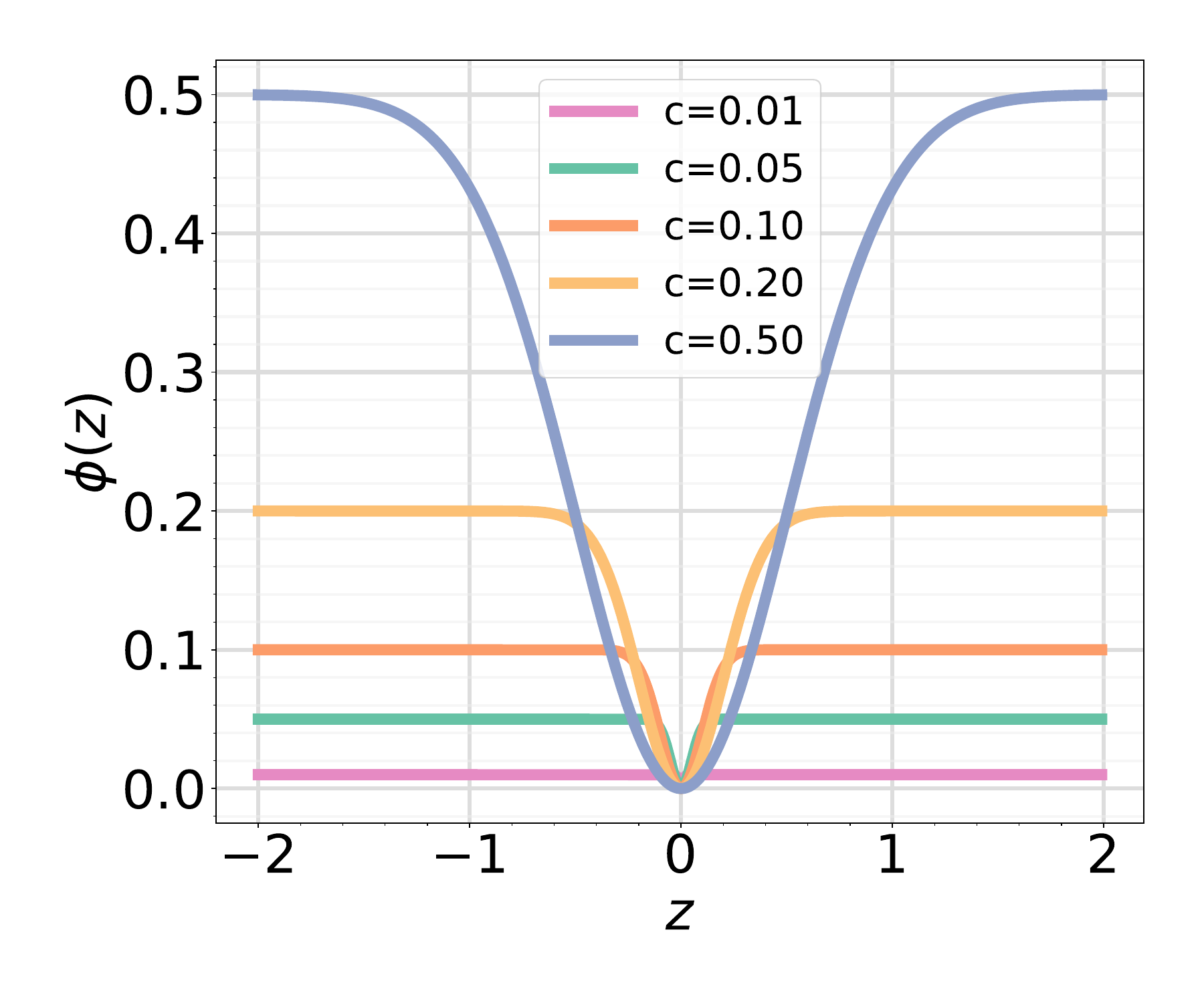}
         \caption{}
         \label{fig:welsch}
     \end{subfigure}
     \begin{subfigure}[b]{0.23\textwidth}
         \centering
         \includegraphics[width=\textwidth]{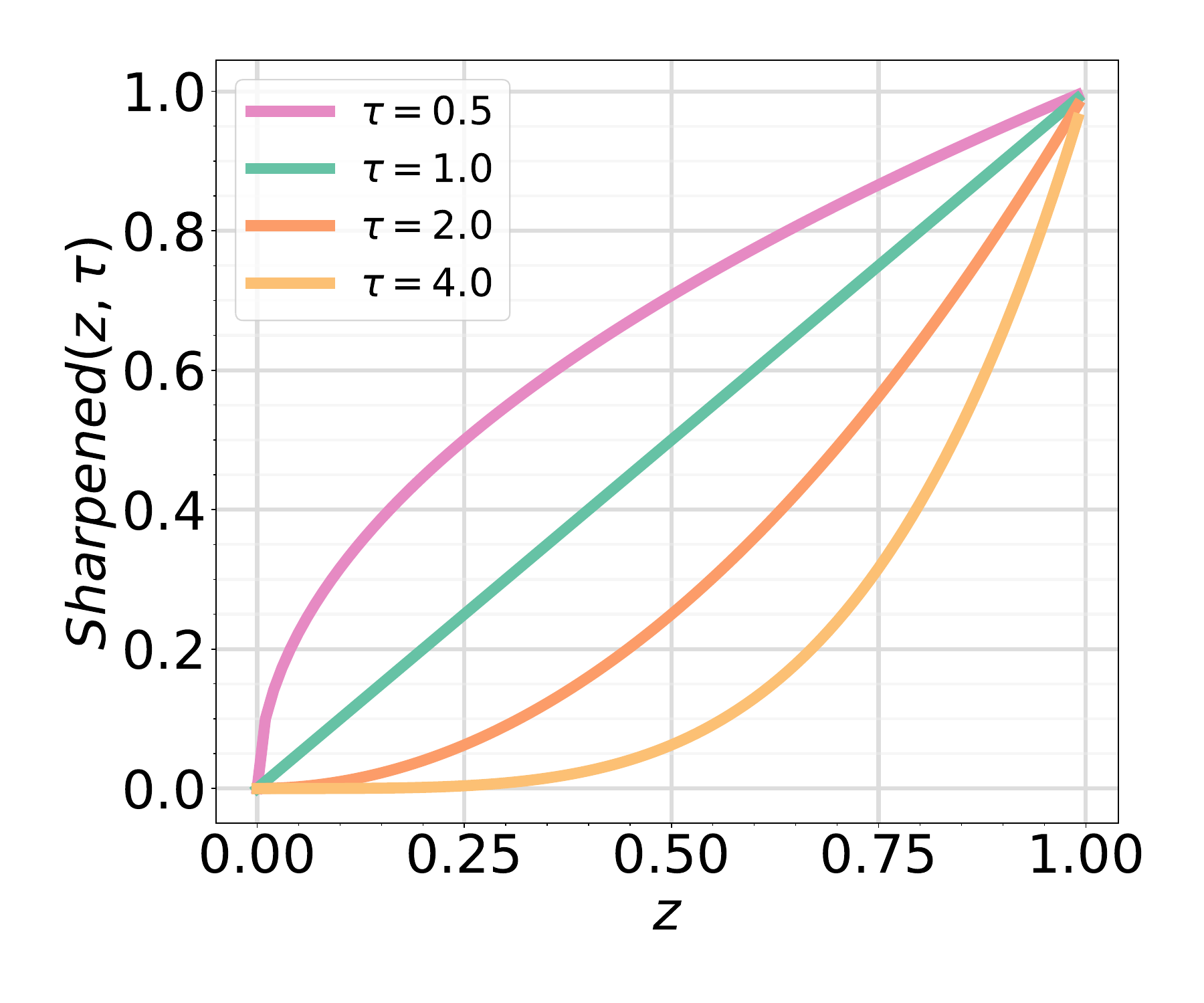}
         \caption{}
         \label{fig:sharpened}
     \end{subfigure}
        \caption{(a) Welsch function with varying $c$. (b) Sharpening the output of $\Upsilon(\mathbf{z}; \mathbf{w})$ (\cref{eqn:sharpen}). Large values of $\tau$ exaggerate small deviation from perfect alignment between $\mathbf{z}$ and $\mathbf{w}$.}
        \label{fig:welsch_sharp}
\end{figure}

\subsection{Robust \ours}
The \textit{\ours} is sensitive to outliers in the input patch~\cite{devlin1975robust}. This can lead to issues when the input distribution changes unpredictably and introduces outliers. For instance, salt and pepper noise can cause large variations in the input energy (first term in the denominator of \Cref{eqn:ncc_first1}) and affect the output of \textit{\ours}. To overcome this, we propose a robust version of the operator called \textit{R-\ours}, which modifies \Cref{eqn:ncc_first1} as follows:
\begin{align*} 
     \Gamma_{u,v,c}(\mathbf{z}; \mathbf{w}) 
     &{}= \frac{\sum_j \left [\phi ( {z}_{\Tilde{u},\Tilde{v}}^{(j)} - \mu_{\mathbf{z}(\Tilde{u},\Tilde{v})} ) \right ] \left [{w}^{(j)}_c - \mu_{\mathbf{w}_c} \right]}{  \left \| \phi( \mathbf{z}_{\Tilde{u},\Tilde{v}} - \mu_{\mathbf{z}(\Tilde{u},\Tilde{v})} ) \right \| \left \|  \mathbf{w}_c - \mu_{\mathbf{w}_c}   \right \| + \epsilon }. \numberthis \label{eqn:robust_ncc_first1}\\
\end{align*}
Here, $\phi(\cdot)$ can be any robust function such as the Huber, Cauchy (aka Lorentzian), Tukey, or Welsch function. In this work, we use the Welsch function due to its simplicity, which is defined as follows~\cite{dennis1978techniques}:
\begin{equation}
    \phi({z}) = c\left[1 - \exp{\left( \frac{-z^2}{2c^2} \right)} \right]
\end{equation}
Here, $c$ is a learnable parameter that controls the amount of penalty for the outliers. The influence of $c$ is depicted in \Cref{fig:welsch}. During the training process, we adopt an initialization strategy where we set the value of the parameter $c$ to a large value. Subsequently, we update the value of $c$ for each layer by computing the mean of the patchwise standard deviation of the input over the training dataset. Similar to batch normalization, we incorporate a moving average component to enhance the stability and effectiveness of the normalization process.



\subsection{Improving Convergence} \label{sec:method-improve}
To enhance the convergence of \textit{\ours} and \textit{R-\ours}, we incorporate some modifications to their base formulations. We use the notation $\Upsilon \left (\mathbf{z}; \mathbf{w} \right)$ to denote either $\Psi \left (\mathbf{z}; \mathbf{w} \right)$ or $\Gamma \left (\mathbf{z}; \mathbf{w} \right)$ in the following text.

\vspace{.5em}
\noindent \textbf{Sharpening:} 
The peaks and valleys of the output $\Upsilon(\mathbf{z}; \mathbf{w})$ can be emphasized (or de-emphasized) by raising it to power $\tau$:
\begin{equation} \label{eqn:sharpen}
    \widetilde{\Upsilon}_{[1]}(\mathbf{z}; \mathbf{w}) = \left \{ \mathrm{max} \left[ 0, \Upsilon \left (\mathbf{z}; \mathbf{w} \right)  \right] \right \}^\tau
   .
\end{equation}
Here, we solely consider the positive outputs, as our empirical observations indicate that this choice stabilizes the training process and prevents convergence to undesirable optima.
\Cref{fig:sharpened} show the relationship between the input and output of the above operation.
Similar, techniques have also been used in cosine similarity-based techniques~\cite{wu2022exploring}. However, our approach differs from cosine similarity-based methods in that we do not pre-determine the value of $\tau$. Instead, we treat it as a learnable parameter and optimize it alongside the weights.

\vspace{.5em}
\noindent \textbf{Gradient Scaling:}
The weight normalization in \Cref{eqn:robust_ncc_first1} tends to reduce the gradient magnitude, which in turn leads to slower convergence~\cite{liu2018decoupled}.
To mitigate this issue, we propose \textit{gradient scaling} using a learnable scaling factor~$\mathbf{A}$. More specifically, we apply the $\mathbf{A}$ to the output of $\widetilde{\Upsilon}_{[1]}(\mathbf{z}; \mathbf{w})$ at every layer, as shown below: 
\begin{equation}
    \widetilde{\Upsilon}_{[2]}(\mathbf{z}; \mathbf{w}) = \mathbf{A} \odot \widetilde{\Upsilon}_{[1]}(\mathbf{z}; \mathbf{w}).
\end{equation}
Given that scaling the output of a function by a constant is equivalent to augmenting the gradient of the function by that constant, the proposed method effectively addresses the reduced gradient issue and accelerates the convergence process.

\vspace{.5em}
\noindent \textbf{Norm-based Attention Mask (NBAM):}
The input norm $\|\mathbf{\widetilde{z}}\|$ signifies the importance of the local patch within the input. Here, $\mathbf{\widetilde{z}} =  \mathbf{z}_{\Tilde{u},\Tilde{v}} - \mu_{\mathbf{z}(\Tilde{u},\Tilde{v})} $ for \textit{\ours} and $\mathbf{\widetilde{z}} =  \phi( \mathbf{z}_{\Tilde{u},\Tilde{v}} - \mu_{\mathbf{z}(\Tilde{u},\Tilde{v})} ) $ for \textit{R-\ours}. 
Normalizing with $\|\mathbf{\widetilde{z}}\|$ assigns equal importance to all patches, but this may be problematic when $\|\mathbf{\widetilde{z}}\|$ is very small. Such small values indicate low-variation (low-information) input areas that should not be equally weighted as high-information areas, as this may cause spurious matches with templates.


To address this issue, we propose the \textit{Norm-based Attention Mask} technique, which leverages a lightweight single convolution layer with sigmoid activation, denoted as $\psi$. This function learns to dynamically assign importance weights to image patches. Specifically, by taking $\|\mathbf{\widetilde{z}}\|$ as input, $\psi$ learns to generate patch-wise importance weights $m \in [0, 1]$.
Subsequently, the output is redefined using the following computation:
\begin{align*}
\widetilde{\Upsilon}_{[3]}(\mathbf{z}; \mathbf{w}) = {}& \psi \left ( \|\mathbf{\widetilde{z}}\| \right) \odot \widetilde{\Upsilon}_{[2]}(\mathbf{z}; \mathbf{w})\\ 
{}& ~ + (1-\psi \left ( \|\mathbf{\widetilde{z}}\| \right)) \odot \left \{\widetilde{\Upsilon}_{[2]}(\mathbf{z}; \mathbf{w}) \odot \|\mathbf{\widetilde{z}}\| \right \} \numberthis \label{eqn:mask}
\end{align*}
where $\Gamma(\mathbf{z}; \mathbf{w})\odot\|\mathbf{\widetilde{z}}\|$ represents the output without normalization.

To further ensure that each feature is weighted equally, we normalize each channel in the output tensor:
\begin{equation}
    \widetilde{\Upsilon}_{[4]}(\mathbf{z}; \mathbf{w}) = \frac{\widetilde{\Upsilon}_{[3]}(\mathbf{z}; \mathbf{w}) - \mu_{\widetilde{\Upsilon}}}{\sigma_{\widetilde{\Upsilon}}}
\end{equation}
where $\mu_{\widetilde{\Upsilon}}$ and $\sigma_{\widetilde{\Upsilon}}$ are the mean and the variance of $\widetilde{\Upsilon}_{[3]}(\mathbf{z}; \mathbf{w})$ along each channel.




\begin{table}[t]
\centering
\resizebox{0.44\textwidth}{!}{%
\begin{tabular}{l|cccc|c}
\hline
Method    & SVHN  & MNIST-M & SYN   & USPS  & Avg\\ \hline
ERM       & 27.83 & 52.72   & 39.65 & 76.94 & 49.29    \\
CCSA      & 25.89 & 49.29   & 37.31 & 83.72 & 49.05    \\
d-SNE     & 26.22 & 50.98   & 37.83 & \best{93.16} & 52.05    \\
JiGen     & 33.80 & 57.80   & 43.79 & 77.15 & 53.14    \\
ADA       & 35.51 & 60.41   & 45.32 & 77.26 & 54.63    \\
M-ADA     & 42.55 & 67.94   & 48.95 & 78.53 & 59.49    \\
ME-ADA    & 42.56 & 63.27   & 50.39 & 81.04 & 59.32    \\
RandConv  & 57.52 & \second{87.76}   & 62.88 & 83.36 & 72.88    \\
L2D       & 62.86 & 87.30   & 63.72 & 83.97 & 74.46    \\
MetaCNN   & \best{66.50} & \best{88.27}   & 70.66 & \second{89.64} & \best{78.77}   \\ \hline
\ours     & 59.92 & 66.67   & 67.33 & 85.24 & 69.79    \\
R-\ours   & \second{65.42} & 71.00   & \second{71.25} & 89.04 & 74.18    \\
R-\ours$^{+\text{RC}}$ & 64.25 & 83.22 & \best{72.25} & 88.89 & \second{77.15}    \\ \hline
\end{tabular}%
}
\caption{Results of single domain generalization experiments on the Digits-DG benchmark. Our proposed method demonstrates superior performance compared to adversarial augmentation-based approaches while achieving comparable results with the state-of-the-art. Additionally, our method complements existing data augmentation techniques, including Random Convolution~(RC). The \best{best results} are denoted as bold and underlined, while the \second{second best results} are indicated by being underlined.}
\label{tab:sdg-digits}
\end{table}

\section{Experiments}
\label{sec:results}
\subsection{Datasets and Settings}
\noindent \textbf{Digits-DG:} consists of five distinct subsets: MNIST, MNIST-M, SVHN, SYN and USPS. Each subset represents a different domain with variations in writing styles and quality, scales, backgrounds, and strokes. We mainly utilise the Digits-DG benchmark for single-source domain evaluation and ablation studies. Following~\cite{volpi2018generalizing, qiao2020learning, wang2021learning, wan2022meta}, we chose the first $10,000$ images from both the MNIST training and validation sets as the source dataset.

\noindent \textbf{CiFAR-10-C~\cite{hendrycks2019robustness}:} is typically used for corruption robustness benchmarking. It contains 15 different corruption types that mimic real-world scenarios, such as noise, blur, weather, and digital artifacts. Each corruption type has five levels of severity. We follow the setup of \cite{wan2022meta, wang2021learning} and use the CIFAR-10 training set as the source dataset while images in CIFAR-10-C are used for evaluation.

\noindent \textbf{Camelyon-17-Wilds~\cite{wilds2021}:} comprises 455k histopathology image patches extracted from 1000 whole-slide images (WSIs) of sentinel lymph nodes~\cite{camelyon17}. These WSIs were obtained from five distinct medical centres, with each centre representing a unique domain within the dataset. The primary objective of this dataset is to classify input patches and determine whether the central region contains any tumour tissue. In our experimental setup, we selected each of the domains as the source domain in turn and used the rest of the domains as target domains.

\noindent \textbf{Implementation Details:} Complete details regarding the experimental setup, including the network architecture and model selection, can be found in the supplementary document, providing a comprehensive understanding of our methodology.


\begin{table}[t]
\centering
\resizebox{0.44\textwidth}{!}{%
\begin{tabular}{l|cccc|c}
\hline
Method    & Weather & Blur  & Noise       & Digits & Average \\ \hline
ERM       & 67.28   & 56.73 & 30.02       & 62.30  & 54.08   \\
ADA       & 72.67   & 67.04 & 39.97       & 66.62  & 61.58   \\
M-ADA     & 75.54   & 63.76 & 54.21       & 65.10  & 64.65   \\
ME-ADA    & 74.44   & 71.37 & 66.47       & 70.83  & 70.78   \\
RandConv* & 74.90   & 74.95 & 55.71       & 76.80  & 70.59   \\
L2D       & 75.98   & 69.16 & \second{73.29} & 72.02  & 72.61   \\
MetaCNN       & {\best{77.44}} & \second{76.80}          & {\best{78.23}} & \second{81.26}          & {\best{78.43}} \\ \hline
XCNorm    & 72.98   & 73.76 & 48.15       & 77.52  & 68.10   \\
R-XCNorm  & 75.57   & 75.72 & 60.81       & 77.61  & 72.43   \\
R-XCNorm$^{+\text{RC}}$ & \second{76.99}          & {\best{79.61}} & 63.95                & {\best{82.27}} & \second{75.70}          \\ \hline
\end{tabular}%
}
\caption{Experiments of single domain generalization on CiFAR-10 classification. Models are trained on the CIFAR-10 train set and evaluated on the CIFAR-10-C benchmark. ``*'' represents our implementation of existing work. Our method is also complementary to existing data augmentation methods such as Random Convolution~(RC). The \best{best results} are denoted as bold and underlined, while the \second{second best results} are indicated by being underlined.}
\label{tab:sdg-cifar}
\end{table}

\subsection{Comparisons on Digits-DG}

\noindent \textbf{Results}: Table~\ref{tab:sdg-digits} provides a comprehensive comparison of the out-of-domain generalization performance between our proposed method and state-of-the-art approaches. The results highlight the significant improvements achieved by our proposed method. Both our base method (\ours) and robust variant (R-\ours) showcase substantial enhancements over the ERM baseline ($49.3\%\rightarrow69.8\%$ and $49.3\%\rightarrow74.2\%$, respectively), without the need for data augmentation techniques or extensive network modifications. Notably, our methods outperform adversarial augmentation-based domain generalization approaches, including ADA, M-ADA, and ME-ADA, by a considerable margin. Also, Table~\ref{tab:sdg-digits} highlights the complementary nature of our method with data augmentation techniques, such as Random Convolution (RC)~\cite{xu2021robust}. The combined approach achieves even higher performance, positioning it competitively alongside the complex state-of-the-art method, MetaCNN.


\subsection{Comparisons on CiFAR-10-C}

\noindent \textbf{Results}: The average accuracy for four categories of level-5 severity corruptions is presented in \Cref{tab:sdg-cifar}. Our proposed method showcases substantial improvements over the baseline model (ERM), achieving an impressive accuracy boost from $54.08\%$ to $68.10\%$ without relying on data augmentation techniques. Notably, our approach demonstrates a remarkable $17.03\%$ improvement for \textit{blur} corruptions and a noteworthy $15.22\%$ improvement for \textit{digits} corruptions. Furthermore, our R-\ours~method effectively enhances model robustness against noise corruption, elevating the accuracy from $48.15\%$ to $60.81\%$. It is worth highlighting that our R-\ours~outperforms most of the data augmentation-based approaches by a respectable margin. Additionally, when combined with RC~\cite{xu2021robust}, our method exhibits exceptional performance across all four categories and demonstrates competitive results compared to the current state-of-the-art method.

\subsection{Comparisons on Camelyon-17}
\begin{figure}[t]
    \centering
    \includegraphics[width=0.5\textwidth]{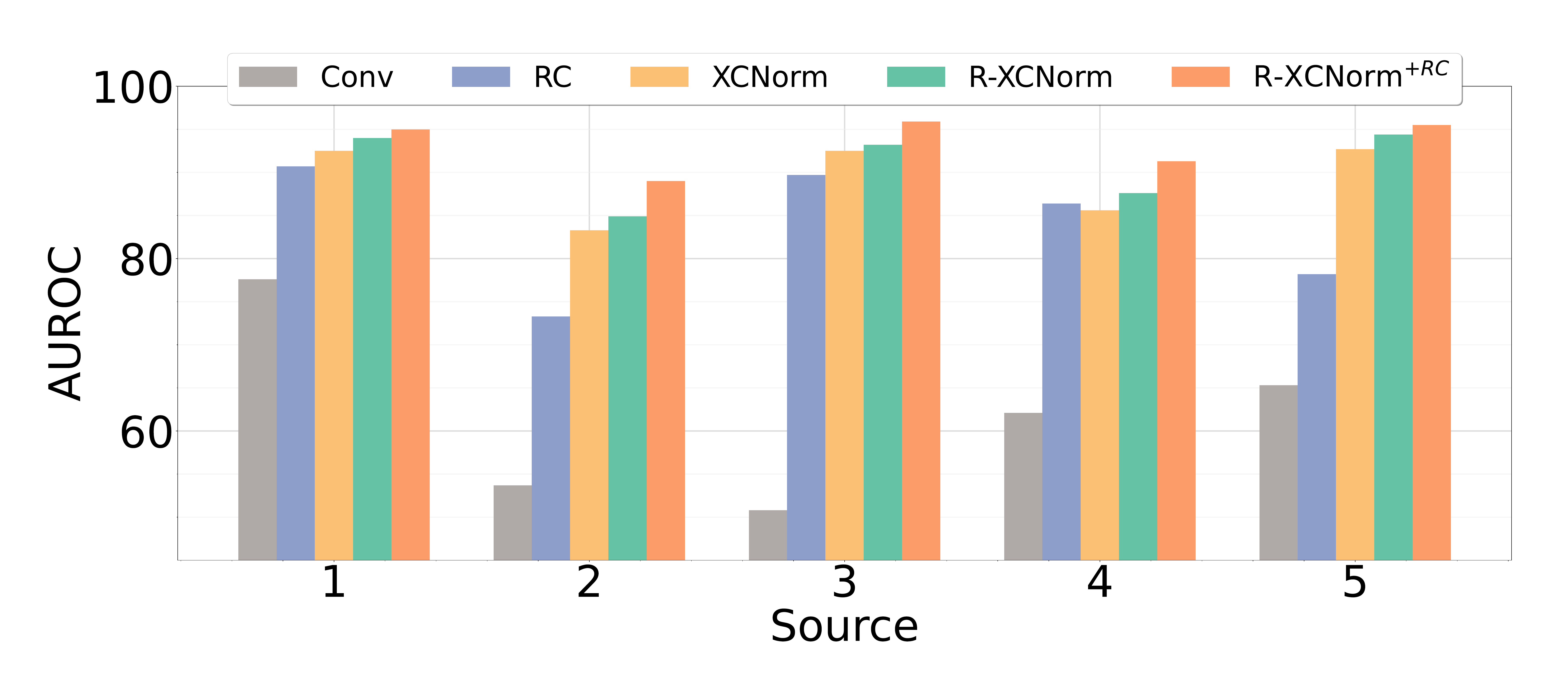}
    \caption{Comparison of model performance on the Camelyon-17 dataset for single domain generalization. Our proposed method surpasses the baseline ERM method, while integrating with the Random Convolution (RC) approach leads to further improvements in domain generalization performance.}
    \label{fig:camelyon17}
\end{figure}

\begin{figure*}
    \centering
    \includegraphics[width=0.9\textwidth]{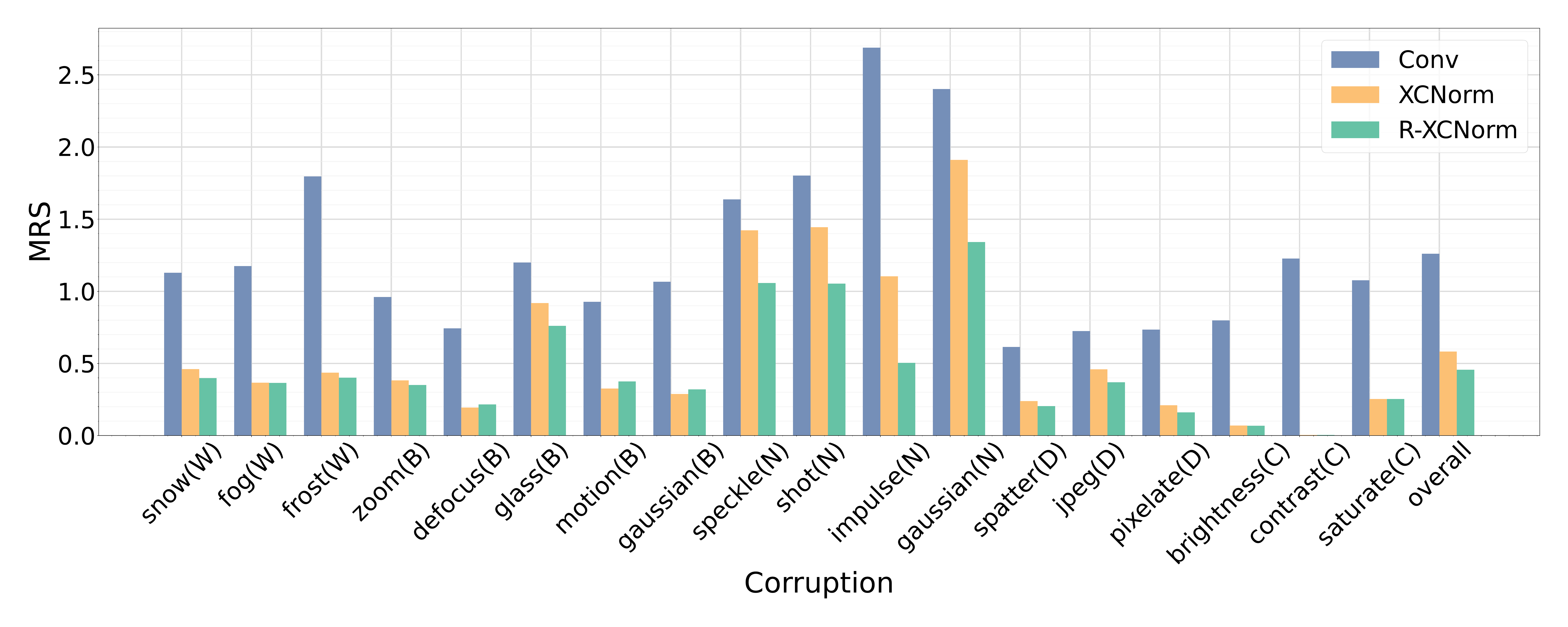}
    \caption{Model Robustness Score (MRS) comparison of different models on the CiFAR-10-C dataset. The MRS values were computed for various categories, including weather (W), blur (B), digital compression (D), noise (N), and chromatic (C) augmentations. Lower MRS values indicate better model robustness against corruption. The baseline ERM model exhibits poor performance, while our proposed method demonstrates significantly better robustness compared to ERM.}
    \label{fig:robustness}
\end{figure*}


\noindent \textbf{Results}: While the Camelyon17 dataset is commonly employed for conventional domain generalization tasks (generalizing from multiple source domains to a single target domain), it has not been extensively explored for single-source domain generalization. In this study, we investigate the single domain generalization performance on the Camelyon17 dataset using the AUROC metric, as shown in \Cref{fig:camelyon17}. Notably, the ERM model exhibits poor generalization when trained on a single source domain, particularly in domains 2 and 3. We attribute this observation to significant variations in the staining agent's color across different hospitals (refer to the supplementary document for examples of training images from different domains). In contrast, the Random Convolution (RC) approach demonstrates impressive domain generalization capabilities on the Camelyon17 dataset.

Remarkably, our proposed \ours~method and its robust variant, R-\ours, consistently outperform the ERM baseline across all domains, without relying on data augmentation techniques. In fact, our methods even surpass the performance of the RC method. Moreover, when combined with RC, our approach achieves a robust model with exceptional domain generalization performance. These results highlight the effectiveness of our method in enhancing domain generalization and its potential to improve model robustness in practical applications, such as medical imaging classification.

\begin{table}[t]
\centering
\resizebox{0.48\textwidth}{!}{%
\begin{tabular}{ccc|cccc|c}
\hline
XCNorm & NBAM & Robust & SVHN & MNIST-M & SYN  & USPS & Avg \\ \hline
\xmark    &  \textbf{--}       & \textbf{--}      & 30.0 & 55.2    & 39.4 & 75.9 & 50.1     \\ \hline
\cmark       &\xmark       & \xmark      & 59.9 & 66.7    & 67.3 & 85.2 & 69.8     \\
\cmark       &\cmark       & \xmark      & 62.0 & 67.4    & 68.8 & 87.9 & 71.5     \\
\cmark       &\xmark       & \cmark      & 64.0 & 69.1    & 70.8 & 87.3 & 72.8     \\
\cmark       &\cmark       & \cmark      & 65.4 & 71.0    & 71.3 & 89.0 & 74.2     \\ \hline
\end{tabular}%
}
\caption{Ablation results on Digits-DG benchmark, and the Top-1 accuracy is reported. Our method can substantially improve the generalization performance of the baseline model.}
\label{tab:ablation-digits}
\end{table}

\section{Discussion}

\subsection{Ablation Study}
In this section, we present the results of our ablation study conducted on the Digits-DG benchmark to evaluate the efficacy of each component of our proposed method. Specifically, we examine the effectiveness of our \ours~method, the robust variant (R-\ours), and the norm-based attention mask (NBAM) proposed to improve model convergence, as discussed in \Cref{sec:method-improve}.

\Cref{tab:ablation-digits} reports the classification results of the four variants of our original framework, including the baseline ERM model for comparison. Our \ours~method, without any extension, achieves a significant $19.7\%$ performance improvement over the baseline, demonstrating the effectiveness of our approach. Furthermore, the integration of NBAM, which relaxes the normalization of input features, leads to additional performance gains. This highlights the importance of selectively normalizing important input regions, as a global normalization approach may mistakenly assign equal importance to insignificant regions and hinder training.

Furthermore, we evaluate the effectiveness of the robust variant, R-\ours, specifically designed for outlier rejection. As illustrated in \Cref{tab:ablation-digits}, the integration of the Welsch robust function enhances performance across all domains except for the USPS dataset. Moreover, combining the robust variant with NBAM yields even greater improvements across all domains without sacrificing performance. These results highlight the benefits of incorporating the robust variant and NBAM in our framework, demonstrating their potential for enhancing model performance and domain generalization.

\subsection{Gradient Scaling}
\begin{figure}
    \centering
    \includegraphics[width=0.36\textwidth]{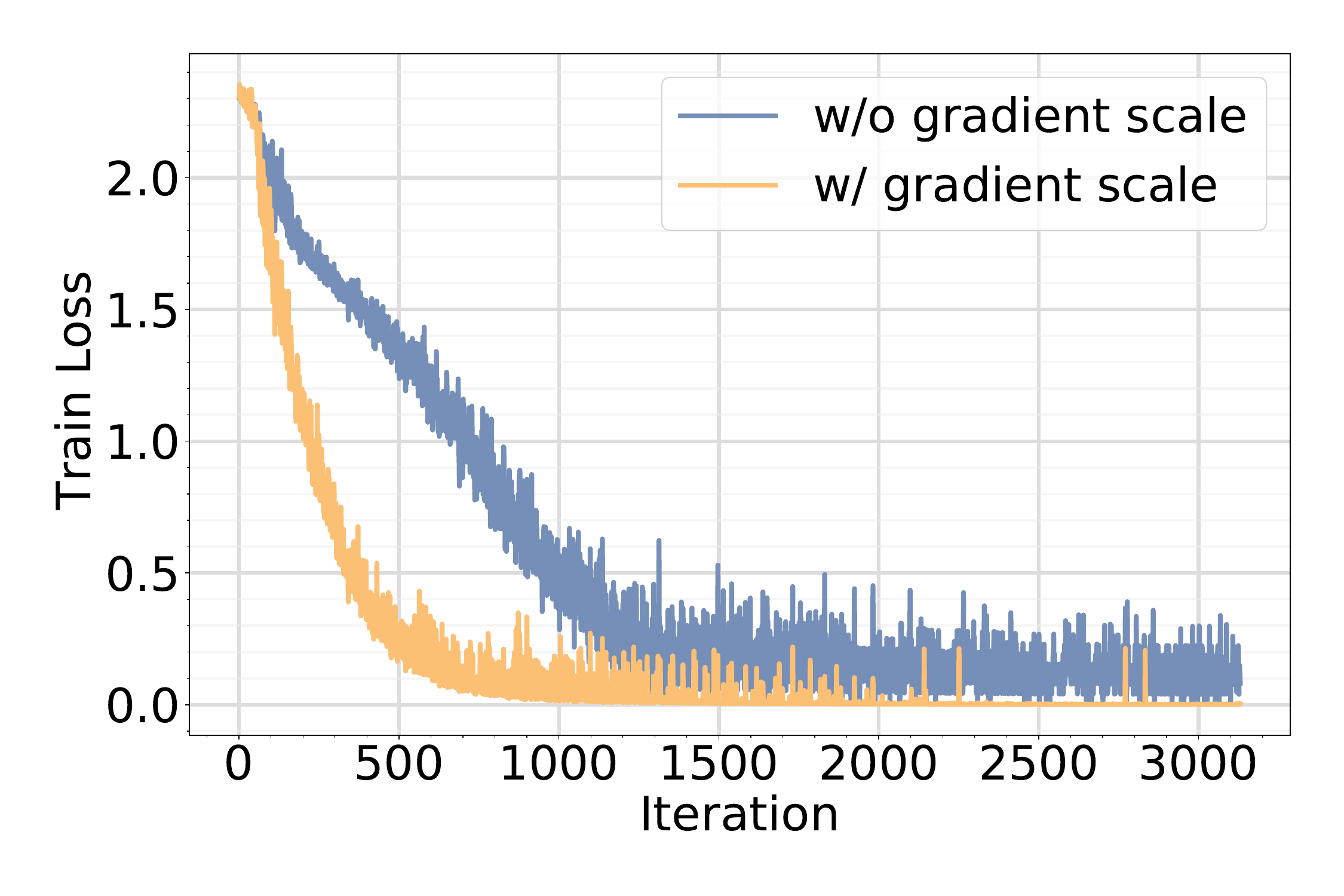}
    \caption{Comparison of optimization convergence rates with and without the proposed gradient scaling. The results highlight the superior convergence speed achieved by incorporating our proposed technique, leading to reduced training time and convergence to a lower optimal point.}
    \label{fig:grad-scale}
\end{figure}

In this section, we investigate the effect of gradient scaling, as proposed in \Cref{sec:method-improve}, on the performance of \ours. As mentioned, The weight normalization employed in \Cref{{eqn:robust_ncc_first1}} tends to reduce the gradient magnitude during backpropagation, as shown in \Cref{eqn:grad-discuss}: 
\begin{equation} \label{eqn:grad-discuss}
    \frac{\partial}{\partial \Tilde{w}}\Gamma(\Tilde{z};\Tilde{w})=\frac{\hat{z}-\hat{w}^\top\hat{z}\cdot\hat{w}}{\|\hat{w}\|}
\end{equation} 
where $\hat{z}=\Tilde{z}/|\Tilde{z}|$ and $\hat{w}=\Tilde{w}/|\Tilde{w}|$. From \Cref{eqn:grad-discuss}, we observe that the large norm $|\hat{w}|$ can diminish the gradients, significantly slowing down the convergence of the optimization process.

To address this issue, we propose a gradient scaling method. As shown in \Cref{fig:grad-scale}, we compare the convergence of \ours~with and without the proposed gradient scaling. The results clearly demonstrate that \ours~with gradient scaling achieves faster convergence than the variant without gradient scaling. This confirms the effectiveness of our proposed gradient scaling method in overcoming the problem of diminishing gradients caused by weight normalization, ultimately speeding up the training convergence.

\subsection{Model Sensitivity to Input Perturbations}
In this section, we investigate the sensitivity of our model to input perturbations using the CiFAR-10-C dataset. Our analysis focuses on the variance in predicted class probabilities, $P(y|x)$, when using a set of corrupted images with different severity levels (denoted as $s$) ranging from 0 to 5. A severity level of 0 represents no corruption, while a severity level of 5 indicates the most severe corruption. 

As depicted in \Cref{fig:dataset_shift}, the input $X$ is influenced by the causal factor $X_0$ (e.g., semantics) and the rendering factor $R$, which contributes to domain shift. Since the objective of S-DG is to develop a model that is invariant to $R$ as much as possible, it is crucial to examine the effect of $R$ on the predictions of the developed model. We measure this by computing the Model Robustness Score (MRS), which quantifies the discrepancy between the predictions obtained using clean and perturbed inputs:
\begin{equation}
\text{MRS}(x, \xi) = \frac{1}{5}\sum_{s=1}^5 KL\left( f_\theta \left( \xi \left( x;0 \right) \right) ; f_\theta\left(\xi\left(x;s\right)\right)\right)
\end{equation}
where, $\xi(x;s)$ represents the perturbation (e.g., blur, noise, compression, weather) included in the CiFAR-10-Corruption dataset. A model that relies solely on $X_0$ for prediction would exhibit a lower MRS compared to a model that incorporates $R$ in its predictions.

\Cref{fig:robustness} shows that the ERM model is highly susceptible to $R$, leading to large MRS values across all categories. In contrast, the proposed \ours~effectively improves the model's robustness, resulting in models that are less reliant on $R$ for prediction. Specifically, our method significantly reduces MRS for the weather, blur, and compression categories. Moreover, our robust variant R-\ours~further diminishes the impact of $R$ on the model's predictions, particularly in the noise category, thereby enhancing both the model's robustness and domain generalization performance. These findings highlight the effectiveness of our approach in mitigating the influence of domain-specific factors and improving the model's sensitivity to the underlying semantics $X_0$ rather than the rendering factor $R$.

\section{Conclusion}
In this paper, we introduce XCNorm, a novel normalization technique based on normalized cross-correlation. XCNorm exhibits invariance to affine shifts and changes in energy within local feature patches, while also eliminating the need for non-linear activation functions. The robust variant, R-XCNorm, focuses on outlier rejection, resulting in improved performance in challenging domains while maintaining competitiveness in others. Our proposed masking method selectively normalizes important input regions, enhancing model stability and out-of-domain performance. The integration of both methods showcases their complementary nature, leading to further improvements across all domains. We demonstrate the practical applicability of XCNorm in medical imaging classification, where it enhances model robustness and sensitivity to underlying semantics. Overall, our work provides effective methods for enhancing model performance and robustness, making notable contributions to the field of single-source domain generalization. These contributions have potential implications for various real-world applications.


{\small
\bibliographystyle{ieee_fullname}
\bibliography{egbib}
}

\end{document}